\documentclass[letterpaper, 10 pt, conference]{ieeeconf}  % Comment this line out if you need a4paper

\IEEEoverridecommandlockouts                              % This command is only needed if 
\let\labelindent\relax
\usepackage[mode=buildnew]{standalone}
\usepackage{graphicx} % Required for including pictures
\usepackage{float}    % For tables and other floats
\usepackage{verbatim} % For comments and other
\usepackage{amsmath}  % For math
\usepackage{amssymb}  % For more math
\usepackage{hyperref}
\usepackage{listings} % For source code
\usepackage{subfig}   % For subfigures
\usepackage{enumitem} % useful for itemization
\usepackage{siunitx}  % standardization of si units
\usepackage[font=footnotesize]{caption}
\usepackage[ruled,vlined,linesnumbered]{algorithm2e}
\usepackage{algpseudocode}
\usepackage{multirow}
\usepackage{array}
\usepackage{booktabs}
\usepackage{gensymb}

\usepackage[T1]{fontenc}
\usepackage{tikz,bm} % Useful for drawing plots
\usepackage{pgfplots}
\usetikzlibrary{positioning}
\pgfplotsset{plot coordinates/math parser=false}
\usetikzlibrary{pgfplots.groupplots}
\usetikzlibrary{external}
\usetikzlibrary{matrix,positioning}
\usepackage{tikzscale}
\usepackage{graphics} % for pdf, bitmapped graphics fsiles
\usepackage{times} % assumes new font selection scheme installed
\usepackage{xcolor}
\usepackage{calc}

\usepackage[compact]{titlesec}
\titlespacing{\section}{0pt}{0.75ex}{0.75ex}
\titlespacing{\subsection}{0pt}{0.5ex}{0.5ex}

\title{\LARGE \bf
A Gaussian Process Model for Opponent Prediction in Autonomous Racing
}

\author{Edward L. Zhu, Finn Lukas Busch, Jake Johnson, and Francesco Borrelli% <-this % stops a space
\thanks{This work was supported by the National Science Foundation under Grant No. 1931853.}% <-this % stops a space
\thanks{Edward L. Zhu, Finn Lukas Busch, Jake Johnson, and Francesco Borrelli {\{\tt\small edward.zhu, finn.busch, jakestrj, fborrelli\}@berkeley.edu} are with the Department of Mechanical Engineering at the University of California, Berkeley, Berkeley, CA 94720, USA}% <-this % stops a space
}%

\begin{document}

\maketitle
\thispagestyle{empty}
\pagestyle{empty}

%%%%%%%%%%%%%%%%%%%%%%%%%%%%%%%%%%%%%%%%%%%%%%%%%%%%%%%%%%%%%%%%%%%%%%%%%%%%%%%%
\begin{abstract}
In head-to-head racing, an accurate model of interactive behavior of the opposing target vehicle (TV) is required to perform tightly constrained, but highly rewarding maneuvers such as overtaking. However, such information is not typically made available in competitive scenarios, we therefore propose to construct a prediction and uncertainty model given data of the TV from previous races. In particular, a one-step Gaussian process (GP) model is trained on closed-loop interaction data to learn the behavior of a TV driven by an unknown policy. Predictions of the nominal trajectory and associated uncertainty are rolled out via a sampling-based approach and are used in a model predictive control (MPC) policy for the ego vehicle in order to intelligently trade-off between safety and performance when attempting overtaking maneuvers against a TV. We demonstrate the GP-based predictor in closed loop with the MPC policy in simulation races and compare its performance against several predictors from literature. In a Monte Carlo study, we observe that the GP-based predictor achieves similar win rates while maintaining safety in up to 3x more races. We finally demonstrate the prediction and control framework in real-time in a experimental study on a 1/10th scale racecar platform operating at speeds of around 2.8 m/s, and show a significant level of improvement when using the GP-based predictor over a baseline MPC predictor. Videos of the hardware experiments can be found at \url{https://youtu.be/KMSs4ofDfIs}.
% and our code has been made available at \url{https://github.com}.
\end{abstract}

%%%%%%%%%%%%%%%%%%%%%%%%%%%%%%%%%%%%%%%%%%%%%%%%%%%%%%%%%%%%%%%%%%%%%%%%%%%%%%%%
\section{Introduction}
A major challenge in automated vehicles is the modeling of interactions between agents in highly dynamic constrained environments. Knowledge of such interactions is crucial for short-term decision making and can have a significant impact on the vehicle's safety and performance. This is especially relevant in situations where information is not shared between agents and inference must be performed to obtain a prediction of future plans of the other agents in the environment. 
In this work, we are interested in the context of racing where agents are each driven by competitive and possibly adversarial policies. Maximizing performance in racing typically requires the agent to operate in highly dynamic regimes, e.g. at the limits of tire friction, where good models are vital to maintaining stability of the vehicle. At the same time, as it is unlikely that a perfect interaction model can be obtained, it is crucial to reason about the uncertainty of the generated predictions, which can help maintain the delicate balance between safety and performance.

Obtaining exact knowledge of an opponent's dynamics and its racing policy is unlikely. However, it is certainly possible that data about them is available from past races. We therefore propose a Gaussian process (GP) based prediction model which learns the \emph{one-step} closed-loop behavior of an opponent, or target (TV), vehicle from a database of interactions and obtains trajectory predictions and the associated uncertainties via sampling. In order to show the benefits of the proposed approach, we formulate a model predictive control (MPC) policy for the ego vehicle (EV) which leverages the time-varying uncertainties provided by the GP-based predictor by constructing uncertainty-expanded collision avoidance constraints to balance performance and safety in a head-to-head racing scenario. We demonstrate through a Monte Carlo simulation study that the GP-based predictor achieves similar win rates while maintaining safety in up to 3x more races when compared to predictors in literature which are based on optimal control and vehicle dynamics. We additionally demonstrate the prediction and control framework in real-time in a experimental study and show a significant level of improvement over a baseline predictor. This work expands upon a previously presented workshop paper with the hardware study to show the feasibility of our approach on a real 1/10th scale racecar platform operating at high speeds of around 2.8 m/s.

\subsection{Related Work}

Recent research in the construction of driver prediction models has focused on utilizing learning-based methods for short-term trajectory prediction. Notably, methods such as \cite{hong2019rules}, \cite{djuric2020uncertaintyaware}, and \cite{WaymoZhao2020TNTTT} infer high level human intent, predicting agent trajectories based on goal states given a semantic map. With respect to prediction under uncertainty, \cite{hong2019rules} models uncertainty in other road agents with a bivariate Gaussian over every future position, whilst considering short-term agent history. Much of the above work takes a model-free approach to trajectory forecasting by embedding interactions into a surrounding "scene context". Alternatively, \cite{Salzmann2020TrajectronDT} proposes a graph-based prediction method that incorporates agent dynamic constraints and \cite{UberCui2020DeepKM} explicitly embeds kinematic feasibility into the last layer of the prediction model. While these deep learning based methods can certainly be applied in the context of racing, their use of semantic images as model inputs leads to the requirement of large motion datasets for training. In contrast, our approach leverages closed-loop state trajectories and can achieve good performance after training on a relatively small dataset of 5000 data points. 

There also exists much work which incorporates learned prediction models in closed-loop with optimal control.  \cite{Vallon2022DataDrivenSF} maps features of the environment to strategy states through a GP, which are then used to construct MPC terminal constraints. This approach is complementary to our proposed method in the sense that it predicts the free space available to the EV under a certain environment configuration instead of the space occupied by the TV. 
\cite{lefevre2015driver} and \cite{Yoon2021InteractionAwarePT} propose learning a mapping from vehicle state history to a finite set of behavioral parameters with a hidden Markov model and GP respectively. In \cite{lefevre2015driver} these parameters are used to modify an MPC reference whereas in \cite{Yoon2021InteractionAwarePT} they are used to construct a measurement model which allows for trajectory predictions of other agents using an extended Kalman filter. Compared to these approaches, we not only condition the TV predictions on the current state of all agents, but also on the future plan of the EV.
\cite{Brdigam2021GaussianPS} is the most similar to our work, where a GP-based state transition model for the TV is learned and the distribution over the predicted trajectory is used in the construction of tightened half-space collision avoidance constraints for a stochastic MPC. The approach is demonstrated on straight track segments for a linearized kinematic vehicle model. Additionally, a "one-move" rule is assumed for the TV, which restricts its competitiveness. In contrast, we do not make the one-move assumption and formulate the approach for a nonlinear dynamic vehicle model and ellipsoidal collision avoidance constraints.

\section{Problem Formulation}

\subsection{Vehicle Model}
We model the racing agents using the dynamic bicycle model \cite{Kong2015}, where the vehicle state and input vectors are defined as $z = [p_x, p_y, \phi, v_x, v_y, \omega]^\top$ and $u = [F_x^r, \delta]^\top$. $p_x$ and $p_y$ are the Cartesian coordinates of the vehicle's center of gravity (CoG), $\phi$ is the vehicle's heading, $v_x$ and $v_y$ are the longitudinal and lateral velocity of the CoG respectively, and $\omega$ is the yaw rate. The control inputs to the vehicle are the rear longitudinal tire force $F_x^r$ and the steering angle $\delta$. We discretize the nonlinear dynamics using the 4\textsuperscript{th} order Runge-Kutta method with time step $T_s$ to obtain:
\begin{align} \label{eq:dynamic_bike_model}
    z_{k+1} = f(z_k, u_k).
\end{align}

% Though we do not model its evolution in \eqref{eq:dynamic_bike_model}, we 
We additionally make use of the vehicle pose in a curvilinear reference frame w.r.t. the centerline of a track \cite{micaelli1993trajectory}, which is represented via the twice continuously differentiable mapping $\tau:[0, L] \mapsto \mathbb{R}^2$, where $L$ is the length of the track and $\tau(0) = \tau(L)$. We denote the first and second derivatives of $\tau$ w.r.t. the track progress $s$ as $\tau',\tau'':[0, L] \mapsto \mathbb{R}^2$. 
% As an example, consider a circular track with radius $r$. Then we have $L=2\pi r$, $\tau(s) = [r\cos(2\pi s/L), r\sin(2\pi s/L)]^\top$, and $\tau'(s) = [-\sin(2\pi s/L), \cos(2\pi s/L)]^\top$. 
The curvilinear pose can be computed from the global position $p=[p_x,p_y]^{\top}$ and heading angle $\phi$, and consists of the vehicle's track progress $s(p) = \arg \min_s \|\tau(s) - p\|_2$, its lateral deviation from the centerline $e_y(p) = \min_s \|\tau(s) - p\|_2$, and its heading deviation from the centerline tangent angle $e_\phi(p,\phi) = \phi - \arctan(\tau'(s(p)))$. We can additionally define the signed curvature of the track at some $s \in [0,L]$ as $|\kappa(s)| = \|\tau''(s)\|_2$, where the sign of $\kappa$ depends on the sign of the swept angle rate for the track at $s$. 
We distinguish between quantities for the EV and TV by adding a superscript, e.g. $z^{(1)}$ and $z^{(2)}$ for the state of the EV and TV respectively.

\subsection{Model Predictive Contouring Control (MPCC)}

In a head-to-head competition, it would be incorrect to assume that knowledge of the opponent's plans or their racing control policy are freely available. We therefore require a prediction model to generate a forecast of the opposing agent's behavior before computing our own control action. Design of such a prediction model is a key contribution of this work and will be described in further detail soon. Let us define the prediction model $\mathcal{P}$ for some horizon of length $N$ as a function of the TV's current state $z_k^{(2)}$, which we assume can be accurately obtained through measurement, a sequence of EV states $\mathbf{z}_k^{(1)} = \{z_k^{(1)}, \dots, z_{k+N}^{(1)}\}$, which describes the EV's planned trajectory over the next $N$ time steps, and hyperparameters $\theta$:
\begin{align} \label{eq:predictor}
    \hat{\mathbf{z}}_k^{(2)} = \mathcal{P}(z_k^{(2)}, \mathbf{z}_k^{(1)}, \theta).
\end{align}

At time step $k$, given a prediction $\hat{\mathbf{z}}_k^{(2)}$ from \eqref{eq:predictor}, we define a baseline optimal control problem for the EV adapted from MPCC \cite{liniger2015optimization}:
\begin{subequations} \label{eq:mpcc}
\begin{align}
    \mathbf{u}^{(1),\star}(&z_k^{(1)},\hat{\mathbf{z}}_k^{(2)}, u_{k-1}^{(1)}) = \nonumber\\
    \arg \min_{\mathbf{u}} \ & \ \sum_{t=0}^{N-1} q_c^{(1)} e_c(p_t, \bar{s}_t)^2 + u_t^\top R^{(1)} u_t \nonumber\\ 
    & \qquad \qquad + \Delta u_t^{\top}R_d^{(1)} \Delta u_t - q_s^{(1)} \bar{s}_N \label{eq:mpcc_cost} \\
    \text{s.t.} \ & \ z_0 = z_k^{(1)} \label{eq:mpcc_init_z} \\
    & \ z_{t+1} = f^{(1)}(z_t, u_t), \ t = 0, \dots, N-1 \label{eq:mpcc_dyn_z} \\
    & \ \bar{s}_0 = s(p_0) \label{eq:mpcc_init_s} \\
    & \ \bar{s}_{t+1} = \bar{s}_t + T_s v_{x,t}, \ t = 0, \dots, N-1 \label{eq:mpcc_dyn_s} \\
    & \ z_t \in \mathcal{Z}, \ t = 0, \dots, N \label{eq:mpcc_z_constr} \\
    & \ u_t \in \mathcal{U}, \ t = 0, \dots, N-1 \label{eq:mpcc_u_constr} \\
    & \ h(z_t, \hat{z}_{k+t}^{(2)}) \leq 0, \ t = 0, \dots, N, \label{eq:mpcc_obs_constr}
\end{align}
\end{subequations}
where $\mathbf{u} = \{u_0, \dots, u_{N-1}\}$, $\Delta u_t = u_t - u_{t-1}$, and $u_{-1} = u_{k-1}^{(1)}$. The objective \eqref{eq:mpcc_cost} is to maximize progress while following the track. These goals are quantified by $\bar{s}_N$ and $e_c$, which are approximations of track progress and the centerline deviation respectively \cite{liniger2015optimization}, and are weighted by parameters $q_s^{(1)}$, $q_c^{(1)} > 0$. We additionally wish to minimize control amplitude and rate via $R^{(1)}$, $R_d^{(1)} \succ 0$. Vehicle state and approximate track progress evolution are governed by the dynamics in \eqref{eq:mpcc_init_z}-\eqref{eq:mpcc_dyn_s}. We impose track boundary and input constraints via \eqref{eq:mpcc_z_constr} and \eqref{eq:mpcc_u_constr} as follows:
\begin{align}
    \mathcal{Z} &= \{z \ | \ -W/2 \leq e_y(p) \leq W/2\} \\
    \mathcal{U} &= \{u \ | \ u_l \leq u \leq u_u\}.
\end{align}
Where $W$ is the width of the track and $u_l$ and $u_u$ are the lower and upper bounds on the input values. Constraint \eqref{eq:mpcc_obs_constr} describes coupling between the predicted EV and TV trajectories which, for the purposes of this work, represent collision avoidance constraints and will be expanded upon later. Using the optimal open-loop sequence from \eqref{eq:mpcc}, we can define the EV feedback policy:
\begin{align} \label{eq:mpcc_policy}
    \pi^{(1)}(z_k^{(1)},&\hat{\mathbf{z}}_k^{(2)}, u_{k-1}^{(1)}) = u_0^{(1),\star}(z_k^{(1)},\hat{\mathbf{z}}_k^{(2)}, u_{k-1}^{(1)}),
\end{align}
which can be applied in a receding horizon fashion. See \cite{liniger2015optimization} for more details on the formulation of MPCC.

% \subsubsection{Target Vehicle}
%     The target vehicle is modeled using the same dynamic bicycle model as \eqref{eq:dynamic_bike_model}. The target vehicle is controlled by the unknown policy:
% \begin{equation} \label{eq:tv_policy}
%     u_k^{\text{TV}} = \pi_k^{\text{TV}}(z_k^{\text{TV}}, z_k^{\text{EGO}})
% \end{equation}
% The resulting closed-loop dynamics of the target vehicle can be written as
% \begin{equation} \label{eq:tv_cl_dynamics}
%     z_{k+1}^{\text{TV}} = g^{\text{TV}}(z_k^{\text{TV}}, z_k^{\text{EGO}})
% \end{equation}

% \subsection{General MPC Formulation}
% We formulate a discrete time constrained optimal control problem over a receding horizon over $N$ timesteps.

% The evolution of the vehicle is governed by discrete-space dynamics,

% $$
% \bf{z_{k+1}}= f_d(\bf{z_k}, \bf{u_k})
% $$

% discretized with a sampling time $\Delta t$ and forward integrated using a fourth-order Runge-Kutta method.
% \subsubsection{Contouring Control}
% How much should we explain here?
% Lag and contour error terms
% \begin{subequations}
% \begin{align}
% e_k = 
% \begin{bmatrix}
% \tilde{e}^l(z_k, s_k) \\
% \tilde{e}^c(z_k, s_k)
% \end{bmatrix} 
% \end{align}

% \begin{align}
% \tilde{e}^l(z_k, s_k) &= -\cos(\phi(s_k)) (x_k - x^R(s_k)) \\ 
% &\nonumber - \sin(\phi(s_k)) (y_k - y^R(s_k)) \\
% \tilde{e}^c(z_k, s_k) &= \sin(\phi(s_k)) (x_k - x^R(s_k)) \\ 
% &\nonumber - \cos(\phi(s_k)) (y_k - y^R(s_k))
% \end{align}
% \end{subequations}
% \subsubsection{Constraints}

\section{Learning TV Behavior via Gaussian Processes}

In any competitive scenario, information about one's capabilities and strategy are closely guarded. We therefore make few assumptions about knowledge of the TV's dynamics or racing policy and simply define the evolution of the TV's state as follows:
\begin{align} \label{eq:tv_dyn}
    z_{k+1}^{(2)} = f^{(2)}(z_{k}^{(2)}, \pi_k^{(2)}(z_{k}^{(2)},z_k^{(1)})) = g_k(z_{k}^{(2)}, z_k^{(1)}),
\end{align}
where $f^{(2)}$ and $\pi_k^{(2)}$ are unknown, but are assumed to be smooth and $g_k$ denotes the closed-loop dynamics of the TV given some EV state $z_k^{(1)}$. 

We assume that a dataset $\mathcal{D}$ of head-to-head racing interactions, which are generated by a stationary TV policy, i.e. $\pi_k^{(2)} = \pi^{(2)}$, is available. Our approach is to construct a probabilistic one-step model of the TV, which approximates $g_k$ in \eqref{eq:tv_dyn} and captures the associated prediction uncertainty. This would allow us to roll out a TV prediction trajectory at time step $k$ given a TV state $z_k^{(2)}$ and EV plan $\mathbf{z}_k^{(1)}$ while maintaining a measure of the uncertainty of our predictions. This is crucial to balancing safety and performance during the highly dynamic and tightly constrained interactions which are commonplace in racing. We note that in the case of non-stationary policies, which are selected from a finite set (e.g. for overtaking, blocking, station keeping, etc.), prediction models can be constructed for each policy option given that the dataset is indexed by policy.

\subsection{Gaussian Processes}
We use GP regression \cite{williams2006gaussian} to model the unknown one-step closed-loop dynamics of the target vehicle \eqref{eq:tv_dyn}. Consider the dataset constructed from $D$ pairs of observations:
\begin{align}
    \mathcal{D} = \{(x^{[i]}, y^{[i]})\}_{i=1}^D,
\end{align}
where $x^{[i]} \in \mathbb{R}^{n}$ and $y^{[i]} \in \mathbb{R}^{m}$ are generated by the unknown statistical process:
\begin{align}
    y^{[i]} = g(x^{[i]}) + w^{[i]},
\end{align}
with i.i.d. $w^{[i]} \sim \mathcal{N}(0, \text{diag}(\sigma_1, \dots, \sigma_m))$. Assuming independence between output dimensions, GPs maintain a probability distribution over each of the functions $g_j$, $j = 1, \dots, m$, and are completely specified by a mean function $\mu_j(x)$ and a scalar covariance function $k_j(x,x')$ (also called the \textit{kernel}). Without loss of generality, given a zero-mean GP prior on $g_j$, the posterior distribution of $g_j$ is given by the GP:
\begin{subequations}
\begin{align}\label{eq:gp_post}
    g_j(x) &\sim \mathcal{N}(\mu_j(x), \text{Var}_j(x)) \\
    \mu_j(x) &=\mathbf{k}_j(x)^\top (\mathbf{K}_j+\sigma_j^2I_D)^{-1}Y_j\\
    \text{Var}_j(x) &= k_j(x,x) - \mathbf{k}_j(x)^\top (\mathbf{K}_j+\sigma_j^2 I_D)^{-1}\mathbf{k}_j(x)
\end{align}
\end{subequations}
where 
$Y_j = [y_j^{[1]} \dots y_j^{[D]}]^\top$, $[\mathbf{K}_j]_{ab}=k(x^{[a]},x^{[b]})$ for $a,b = 1, \dots, D$, and $\mathbf{k}_j(x)=[k_j(x,x^{[1]})\ \dots \ k_j(x,x^{[D]})]^\top$. 
In this work, we use the Mat\'ern kernel with parameter $\nu = 1.5$ \cite{williams2006gaussian}:
\begin{align*}
    k_j(x,x') &= \\ &\left(1+\frac{\sqrt{3}\|x-x'\|^2}{l_j}\right)\exp\left(-\frac{\sqrt{3}\|x-x'\|^2}{l_j}\right),
\end{align*}
where $l_j$ is a length scale parameter which, together with the prior observation variances $\sigma_j$, can be optimized by maximizing the marginal likelihood of the observations using gradient-based methods.

\subsection{Data Generation and Model Training} \label{sec:data_gen}
We construct the dataset $\mathcal{D}$ from simulation rollouts of EV and TV trajectories drawn from head-to-head races. In each of these rollouts, the EV starts behind TV on a randomly generated track. EV uses the policy defined by \eqref{eq:mpcc} and \eqref{eq:mpcc_policy} and can win the race by executing a successful overtaking maneuver. On the other hand, TV tries to remain ahead of EV by blocking attempted overtakes. In order to encourage blocking behavior, the TV policy is formulated obtained by switching the superscripts in \eqref{eq:mpcc} and \eqref{eq:mpcc_policy}, \textit{removing} the collision avoidance constraints \eqref{eq:mpcc_obs_constr}, and adding to \eqref{eq:mpcc_cost} the following term:
\begin{align} \label{eq:blocking_cost}
    \sum_{t=0}^N q_y^{(2)}\frac{(e_y(p_t^{(2)})-e_y(p_k^{(1)}))^2}{1+(s(p_k^{(2)})-s(p_k^{(1)}))^2}
\end{align}
with $q_y^{(2)} > 0$ which rewards the TV for maintaining the same lateral position on the track as the EV especially when the two vehicles are close in terms of their track progress. Note that $q_y^{(2)}$ can be interpreted as an aggressiveness factor for the TV's blocking attempts. In addition, it is important to note that this cost term only depends on the state of the EV at the current time step $k$ and does not require a prediction of the EV's trajectory over the MPC horizon. As such the overall TV policy is only a function of the EV's state at time step $k$. During these data generation rollouts, the predictions $\hat{\mathbf{z}}^{(2)}_k$ are replaced with the open-loop solution to \eqref{eq:mpcc} for the TV. In other words, the EV knows the true plan of its opponent.

The regression features are constructed using the resulting closed-loop trajectories as follows
\begin{align*}
    x^{[i]} =& [\Delta s_k, \Delta e_{y,k}, e_{y,k}^{(2)}, e_{\phi,k}^{(2)}, v_{x,k}^{(2)}, \omega_k^{(2)}, e_{\phi,k}^{(1)}, v_{x,k}^{(1)}, \kappa_k]^\top,
\end{align*}
where $\Delta s_k = s(p_k^{(1)})-s(p_k^{(2)})$, $\Delta e_{y,k} = e_y(p_k^{(1)})-e_y(p_k^{(2)})$, and $\kappa_k$ is a vector of track curvatures at the look-ahead points $s(p_k^{(2)})+i\delta$, $i=1,\dots,V$ for some chosen $V$ and $\delta > 0$. The regression targets are chosen as the one-step TV state difference, which are also transformed into the curvilinear frame via $y^{[i]} = \mathcal{C}(z_{k+1}^{(2)}) - \mathcal{C}(z_k^{(2)})$.
Evaluation of the one-step predictor can be done with:
\begin{align} \label{eq:gp_eval}
    \hat{z}_{k+1}^{(2)} = \mathcal{C}^{-1}(\mathcal{C}(z_{k}^{(2)}) + \mu(z_{k}^{(2)}, z_{k}^{(1)})),
\end{align}
where $\mu$ is the vector-valued function of posterior means.

Transforming vehicle poses to the curvilinear frame has the effect of reducing the amount of data required to achieve good generalization on a wide variety of tracks. This is crucial for GPs which have complexity $\mathcal{O}(D^3)$ and which rely heavily on a distance metric in the kernel function for training and inference. Consider the case of a GP where inference is performed on a track identical to the one used for training data generation but is rotated arbitrarily. It is easy to see that any vehicle pose on the original track which undergoes the same rotation will have the same coordinate in the new curvilinear frame. However, that is obviously not true for the global pose. Therefore, a model trained on data in the curvilinear frame can be immediately applied to the same track with arbitrary orientation, whereas a global frame alternative would be expected to perform poorly as the spatial relationships learned by the model no longer hold on the rotated track.

The look-ahead curvatures $\kappa_k$ were included in the feature vector to further improve generalization to unseen track configurations and is motivated by look-ahead fixations of human drivers when driving through curved roads \cite{lehtonen2013look}.

\subsection{Trajectory Prediction}

\setlength{\textfloatsep}{0pt}% Remove \textfloatsep
\begin{algorithm}[t!]
    \SetAlgoLined
	\KwIn{$N$, $M$, $z_k^{(2)}$, $\mathbf{z}_k^{(1)}$}
	$z_k^{2,[i]} \leftarrow z_k^{(2)}, \ \forall i = 1, \dots, M$\;
	\For{$i = 1, \dots, M$}{
	    \For{$t = 0, \dots, N-1$}{
	        Sample $\Delta z_{k+t}^{(2),[i]} \sim \mathcal{N}(\mu(z_{k+t}^{(2),[i]},z_{k+t}^{(1)}), \text{Var}(z_{k+t}^{(2),[i]},z_{k+t}^{(1)}))$\;
	        $z_{k+t+1}^{(2),[i]} \leftarrow z_{k+t}^{(2),[i]}+ \Delta z_{k+t}^{(2),[i]}$\;
	    }
	}
	$\hat{z}_{k+t}^{(2)} \leftarrow \frac{1}{M}\sum_{i=1}^M z_{k+t}^{(2),[i]}$\;
	$\hat{\Sigma}_{k+t}^{(2)} \leftarrow \frac{1}{M-1}\sum_{i=1}^M (z_{k+t}^{(2),[i]}-\hat{z}_{k+t}^{(2)})(z_{k+t}^{(2),[i]}-\hat{z}_{k+t}^{(2)})^\top$\;
	\KwOut{$\hat{\mathbf{z}}_{k}^{(2)}$, $\hat{\Sigma}_k^{(2)}$, \dots, $\hat{\Sigma}_{k+N}^{(2)}$}
	\caption{TV trajectory prediction and uncertainty propagation}
	\label{alg:traj_pred}
\end{algorithm}

After obtaining the GP prediction model, we can now construct a trajectory prediction for the TV and the associated uncertainty over a finite horizon of length $N$, as conditioned on a given EV plan. We propose a straight-forward sampling-based approach which estimates a nominal TV state and approximately propagates the uncertainty as described by the one-step GP model. 

The method is presented in Algorithm~\ref{alg:traj_pred} and begins with the initialization of $M$ sample trajectories in line 1. Lines 4 and 5 proceed with evaluation of the one-step GP prediction model for each of the $M$ trajectories at time step $k+t$. Note that we omit the transformations in \eqref{eq:gp_eval} for brevity. The resulting distribution is then sampled for the TV state difference, which is then added on to the current TV state to obtain the new state at time $k+t+1$. This procedure is repeated for $N$ steps. In lines 8 and 9, sample means and covariances are computed over the $M$ trajectories at each time step and returned as the predicted TV trajectory and associated uncertainty. As the trajectories as sampled independently, our procedure is amenable to parallelization.

\subsection{Implementation} \label{sec:gp_implementation}

We train a GP with independent output dimensions using {\tt\small GPyTorch} \cite{gardner2018gpytorch} on simulation rollouts where the TV policy's aggressiveness factor is set to $q_y^{(2)}=200$. To enable real-time inference, this model is trained as a variational approximate GP as described in \cite{Hensman2015ScalableVG} through an inducing point method. This reduces the overall sample complexity to $\mathcal{O}(D\bar{D}^2)$ with $\bar{D} \ll D$ inducing points.  Specifically, we use $\bar{D}=200$ inducing points for a dataset of size $D = 5000$. Executing Algorithm~\ref{alg:traj_pred} with $M=10$ sample trajectories yields an average total run time of 45 ms for a horizon of length $N=10$ on an NVIDIA GeForce RTX 3080.

\begin{figure}
    \centering
    \def\svgscale{0.65}
    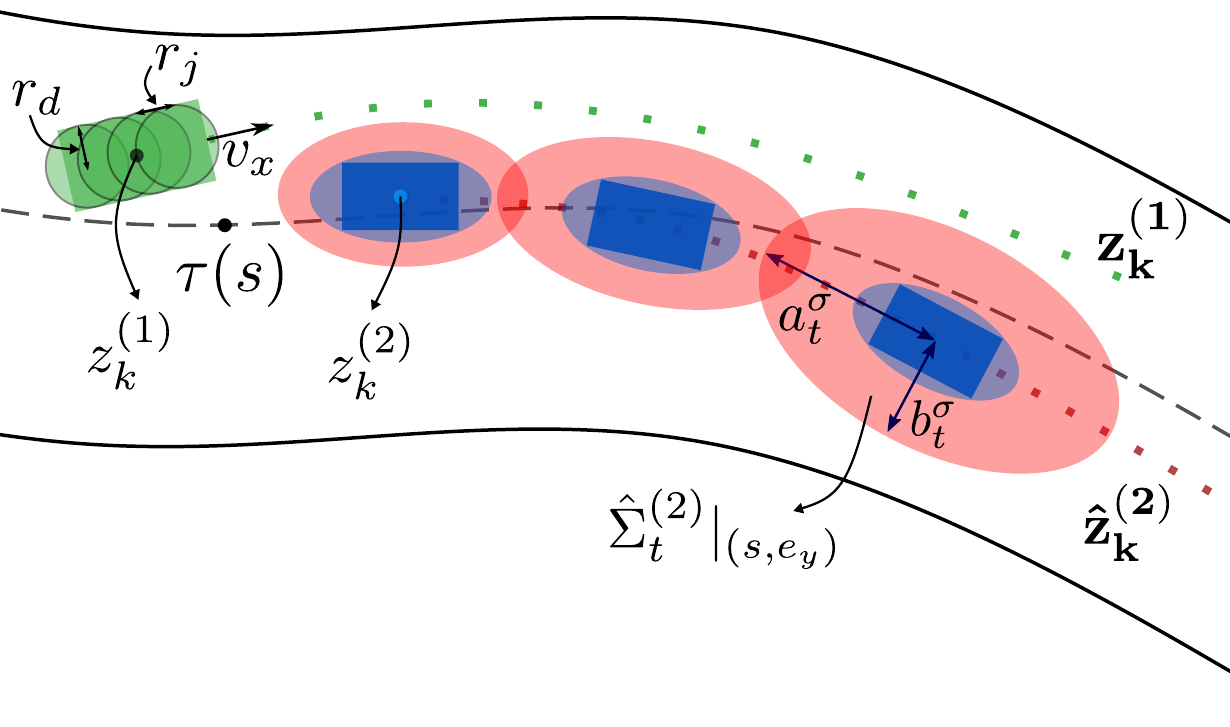
    \vspace{-0.5cm}
    \caption{Overview of the EV obstacle avoidance constraints with uncertainty-expanded ellipses pictured in red and vehicle extent ellipses pictured in blue. The EV is represented by 4 discs of radius $r_d$.}
    \label{fig:overview}
    \vspace{0.1cm}
\end{figure}

\section{Design of MPCC Racing Policy} \label{sec:mpcc_design}

In this section, we introduce modifications to \eqref{eq:mpcc} for the EV, which allow it to leverage the uncertainties provided by Algorithm~\ref{alg:traj_pred} to intelligently trade off between safety and performance. We first define the original obstacle avoidance constraint introduced in \cite{Schwarting2017ParallelAI}, which represents the EV at time step $k$ as a set of four circles with radii $r_d$ and centers $c_{x,k}^{(1),j} = p_{x,k}^{(1)} + r_j\cos(\phi_k^{(1)})$ and $c_{y,k}^{(1),j} = p_{y,k}^{(1)} + r_j\sin(\phi_k^{(1)})$ for user defined coverage constants $r_j$ and $j=1,\dots,4$. The extents of the EV are assumed to be contained within the union of these circles. The TV is represented as the minimum covering ellipse of the vehicle extents, with semi-axis lengths $a$ and $b$, transformed by its pose $(p_k^{(2)}, \phi_k^{(2)})$. These are illustrated in Fig.~\ref{fig:overview} by the green circles and blue ellipses respectively. The obstacle avoidance constraint can be written for $j = 1,\dots, 4$ and $t = k, \dots, k+N$ as
\begin{align} \label{eq:uncert_obs_constraint}
&h_j(z_t^{(1)}, \hat{z}_{t}^{(2)}) = 1\\
&-\frac{((c_{x,t}^{(1),j}-\hat{p}_{x,t}^{(2)})\cos(\hat{\phi}_t^{(2)}) - (c_{y,t}^{(1),j}-\hat{p}_{y,t}^{(2)})\sin(\hat{\phi}_t^{(2)}))^2}{a^2} \nonumber \\
&-\frac{((c_{x,t}^{(1),j}-\hat{p}_{x,t}^{(2)})\sin(\hat{\phi}_t^{(2)}) + (c_{y,t}^{(1),j}-\hat{p}_{y,t}^{(2)})\cos(\hat{\phi}_t^{(2)}))^2}{b^2}\nonumber
\end{align}

We propose an uncertainty-expanded safety bound whose area is proportional to the position prediction variance from Algorithm~\ref{alg:traj_pred}. These are illustrated by the red ellipses in Fig.~\ref{fig:overview}. Let $\text{Var}(s_t^{(2)})$ and $\text{Var}(e_{y,t}^{(2)})$ be the prediction variances for track progress and lateral deviation on the diagonal of $\hat{\Sigma}_{t}^{(2)}$. In order to derive the tightened constraint, we transform these variances from the track-aligned curvilinear frame to the body-aligned frame of the TV:
\begin{align*}
\text{Var}(a_{t}) &= \text{Var}(\cos(\hat{e}_{\phi,t}^{(2)})s_t^{(2)} + \sin(\hat{e}_{\phi,t}^{(2)})e_{y,t}^{(2)}) \\
&=  \cos^2(\hat{e}_{\phi,t}^{(2)})\text{Var}(s_t^{(2)}) + \sin^2(\hat{e}_{\phi,t}^{(2)})\text{Var}(e_{y,t}^{(2)}) \\
\text{Var}(b_{t}) &= \sin^2(\hat{e}_{\phi,t}^{(2)})\text{Var}(s_t^{(2)}) + \cos^2(\hat{e}_{\phi,t}^{(2)})\text{Var}(e_{y,t}^{(2)}),
\end{align*}
where $\text{Var}(a_{t})$ and $\text{Var}(b_{t})$ correspond to the variance in the TV's longitudinal and lateral directions respectively. Note that we neglect off-diagonal terms in $\hat{\Sigma}_{t}^{(2)}$ and the uncertainty in the heading. The semi-axis lengths of the expanded ellipsoidal safety bounds $a_{t}^\sigma$ and $b_{t}^\sigma$ are then computed as
\begin{align} \label{eq:uncertainty_ellipse_semi_axis}
\begin{bmatrix} a_{t}^\sigma \\ b_{t}^\sigma \end{bmatrix} = \gamma \begin{bmatrix} \sqrt{\text{Var}(a_{t})} \\ \sqrt{\text{Var}(b_{t})} \end{bmatrix} (1 - \epsilon_t) + \begin{bmatrix} a \\ b \end{bmatrix},
\end{align}
where $\gamma > 0$ is chosen to be the number of standard deviations we would like to account for in our safety bound, and the constraint function \eqref{eq:mpcc_obs_constr} is defined by replacing $a$ and $b$ in \eqref{eq:uncert_obs_constraint} with the semi-axis lengths computed in \eqref{eq:uncertainty_ellipse_semi_axis}. We allow for a certain amount of constraint violation through the slack variable $\epsilon_t \in [0, 1]$. Note that when $\epsilon_t = 1$, the original constraint \eqref{eq:uncert_obs_constraint} is recovered. We add the following quadratic term to the cost function in \eqref{eq:mpcc_cost}:
\begin{align} \label{eq:slack_cost}
    \frac{1}{2}\epsilon^\top Q_\epsilon \epsilon + q_\epsilon^\top \epsilon,
\end{align}
where $Q_\epsilon \succeq 0$ and $q_\epsilon \geq 0$. This disincentivizes the use of slack variables, i.e. violation of the uncertainty-expanded safety bound, but will allow for a level of risk to be taken when a significant improvement in performance can be achieved. As such, the policy defined by \eqref{eq:mpcc} and \eqref{eq:mpcc_policy} can trade off between safety and performance in a manner dictated by the weights in \eqref{eq:mpcc_cost} and \eqref{eq:slack_cost}.

The nonlinear optimal control problem is formulated using CasADi \cite{Andersson2019} and solved using sequential quadratic approximations with the QP solver \texttt{hpipm} \cite{frison2020hpipm}. Obtaining a solution takes 30 ms on average on a laptop with a 2.6 GHz 9th-Gen Intel Core i7 CPU.

\section{Results and Discussion} \label{sec:results_disc}
% \begin{figure}[t]
%     \centering
%     % \input{plots/track}
%     \includegraphics[width=0.99\columnwidth]{plots/tracks.png}
%     \vspace{-0.5cm}
%     \caption{Tracks used for simulation (left) and hardware (right) experiments.}
%     % \vspace{-0.3cm}
%     \label{fig:tracks}
% \end{figure}
% \begin{figure}[t]
%     \centering
%     \includegraphics[width=0.99\columnwidth]{barc.jpg}
%     \vspace{-0.3cm}
%     \caption{The BARC platform used for hardware experiments.}
%     % \vspace{-0.3cm}
%     \label{fig:barc}
% \end{figure}
Our approach is evaluated in simulation and hardware races, which match the EV, using the MPCC policy from Sec.~\ref{sec:mpcc_design}, against a TV, using the blocking policy defined in Sec.~\ref{sec:data_gen}. Both policies use a horizon length of $N=10$ and sample time of $T_s=0.1$ s. In both simulation and hardware races, the EV starts the race behind the TV and is responsible for avoiding collisions. All crashes during an overtaking attempt are counted as the EV's mistake and consequently result in a loss for the EV. Additionally, leaving the track is also counted as a loss for the EV. Hardware experiments were conducted on the 1/10th scale racecar platform, which are each equipped with Raspberry Pi 4 for onboard computation. Measurement of the vehicle state is done using an OptiTrack motion capture system. The EV and TV policies and predictors run on a host machine and communication with the cars is handled using ROS2. Code to reproduce the simulation results has been made available\footnote{\url{https://github.com/MPC-Berkeley/gp-opponent-prediction-models}}.

To show the importance of predicting interactions between the two vehicles, we take an approach similar to \cite{Yoon2021InteractionAwarePT} and compare the performance of our GP-based approach against a constant velocity (CV) and progress-maximizing nonlinear MPC (NL) predictor in closed-loop with the EV MPCC policy from Sec.~\ref{sec:mpcc_design}. CV propagates the TV's measured state under a constant linear and angular velocity assumption. NL returns the open-loop solution of the optimal control problem defined by removing the collision avoidance constraint \eqref{eq:mpcc_obs_constr} from \eqref{eq:mpcc}. The weights for NL are otherwise chosen to be identical to those used in the TV's blocking policy. It is important to note that NL does \textit{not} include \eqref{eq:blocking_cost} in its cost function. As CV and NL do not come with any measure of uncertainty, we add a constant circular safety bound to encourage fair comparison with the uncertainty ellipses of the GP predictions. The size of this safety bound can be interpreted as a hyperparameter controlling the amount of risk the EV is willing to take in its overtaking attempts when using the CV and NL predictors.

\begin{figure}[t]
  \centering
  \input{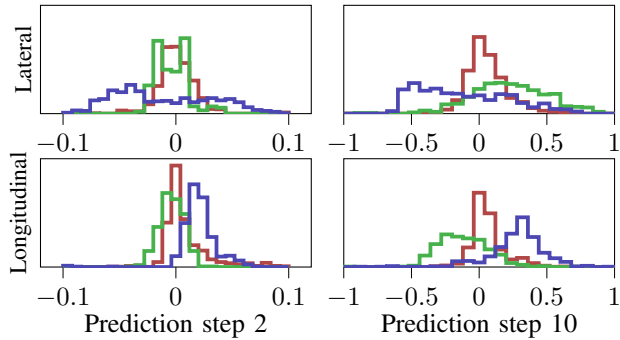}
  \vspace{-0.2cm}
  \caption{Distribution of lateral (top) and longitudinal (bottom) prediction errors [m] at the second (left) and last (right) prediction step for the GP (\textcolor{black!20!red}{\rule[2pt]{10pt}{1pt}}), CV (\textcolor{black!20!green}{\rule[2pt]{10pt}{1pt}}), and NL (\textcolor{black!20!blue}{\rule[2pt]{10pt}{1pt}}) predictors.}
  \label{fig:accs}
  \vspace{-0.3cm}
\end{figure}
\renewcommand{\arraystretch}{1.1}
% \begin{table}[t]
% \centering
% \begin{tabular}{ cc|c|c|c } 
%  & & GP & CV & NL \\
% \midrule[1pt]
% \multirow{2}{*}{Lateral} & mean [m] & \textbf{0.02}  & 0.064 & -0.057\\ 
% & variance [$\text{m}^2$] & \textbf{0.006} & 0.024 & 0.045 \\
% \midrule[0.5pt]
% \multirow{2}{*}{Longitudinal} & mean [m] & \textbf{0.026} & -0.037 &  0.11\\
% & variance [$\text{m}^2$] & \textbf{0.0059} & 0.0094 & 0.026 \\
% \end{tabular}
% \caption{Prediction errors}
% \label{tab:acc}
% % \vspace{-0.7cm}
% \end{table}
\begin{table}[t]
\centering
\begin{tabular}{c|c|c|c} 
 & GP & CV & NL \\
\midrule[1pt]
Lateral & $\mathbf{0.02 \pm 0.077}$  & $0.064 \pm 0.15$ & $-0.057 \pm 0.21$ \\
Longitudinal & $\mathbf{0.026 \pm 0.077}$ & $-0.037 \pm 0.097$ &  $0.11 \pm 0.16$ \\
\end{tabular}
\caption{Prediction errors [m]}
\label{tab:acc}
\end{table}
\subsection{Simulation Study}
In simulation, we perform a Monte Carlo study where 100 starting configurations are sampled and races are run from those initial conditions using each of the aforementioned predictors in close-loop with the EV's policy. For the purposes of the studies below, we vary the value of $\gamma$ from \eqref{eq:uncertainty_ellipse_semi_axis} for GP, the radius of the safety bound for CV and NL, and the blocking aggressiveness parameter $q_y^{(2)}$ of the TV policy.

We first examine the longitudinal and lateral errors of the nominal open-loop predictions with respect to the \emph{actual} trajectory driven by the TV. The errors are defined in terms of track progress $s$ and lateral deviation from the centerline $e_y$ respectively. We restrict our error analysis to predictions made during vehicle interactions, i.e. situations where the two vehicles are within two car lengths of each other. As seen in Table \ref{tab:acc}, GP clearly outperforms both CV and NL, particularly when predicting the blocking behavior of the TV as evidenced by the lower lateral error. This is further illustrated in the prediction error distributions in Fig.~\ref{fig:accs}, where it can be seen that GP maintains a lower bias and variance in both longitudinal and lateral prediction error even at the end of the prediction horizon.

\begin{figure*}[t]
    \centering
    \begin{tikzpicture}
\definecolor{darkgray176}{RGB}{176,176,176}
\definecolor{darkslateblue7671178}{RGB}{76,71,178}
\definecolor{lightgray204}{RGB}{204,204,204}
\definecolor{mediumseagreen7117876}{RGB}{71,178,76}
\definecolor{sienna1787171}{RGB}{178,71,71}
\definecolor{peru17816071}{RGB}{100,100,100}

\begin{groupplot}[group style={group size=2 by 2 , vertical sep=0.2cm, group name=eval, horizontal sep=1.5cm},
width=0.88\columnwidth,
height=0.4\columnwidth]
% Wr Agressive
\nextgroupplot[
legend cell align={left},
legend columns=1,
legend style={fill opacity=0.8, draw opacity=1, text opacity=1, draw=lightgray204,
anchor=west,
at={(2.3,-0.19)}, row sep = 0cm},
tick align=outside,
tick pos=left,
x grid style={darkgray176},
xmin=-0.2, xmax=4.2,
xtick style={color=black},
xtick={0,1,2,3,4},
xticklabels={,,},
y grid style={darkgray176},
ylabel={win rate},
ymin=0.0875, ymax=0.65,
ytick style={color=black},
]
\addplot [semithick, peru17816071,line width=2pt]
table {%
0 0.11
1 0.21
2 0.37
3 0.63
4 0.56
};
\addlegendentry{$GT$}
\addplot [semithick, sienna1787171, mark=*, mark size=2, mark options={solid}]
table {%
0 0.11
1 0.14
2 0.41
3 0.55
4 0.49
};
\addlegendentry{$GP_{0.5}$}
\addplot [semithick, sienna1787171, dotted, mark=triangle*, mark size=2, mark options={solid}]
table {%
0 0.12
1 0.14
2 0.4
3 0.56
4 0.48
};
\addlegendentry{$GP_{1}$}
\addplot [semithick, sienna1787171, dashed, mark=square*, mark size=2, mark options={solid}]
table {%
0 0.12
1 0.12
2 0.41
3 0.51
4 0.47
};
\addlegendentry{$GP_{2}$}
\addplot [semithick, darkslateblue7671178, mark=*, mark size=2, mark options={solid}]
table {%
0 0.12
1 0.18
2 0.17
3 0.22
4 0.19
};
\addlegendentry{$NL_0$}
\addplot [semithick, darkslateblue7671178, dotted, mark=triangle*, mark size=2, mark options={solid}]
table {%
0 0.15
1 0.19
2 0.32
3 0.36
4 0.27
};
\addlegendentry{$NL_{0.025}$}
\addplot [semithick, darkslateblue7671178, dashed, mark=square*, mark size=2, mark options={solid}]
table {%
0 0.15
1 0.21
2 0.33
3 0.33
4 0.31
};
\addlegendentry{$NL_{0.1}$}
\addplot [semithick, mediumseagreen7117876, mark=*, mark size=2, mark options={solid}]
table {%
0 0.14
1 0.19
2 0.33
3 0.49
4 0.55
};
\addlegendentry{$CV_0$}
\addplot [semithick, mediumseagreen7117876, dotted, mark=triangle*, mark size=2, mark options={solid}]
table {%
0 0.13
1 0.14
2 0.27
3 0.45
4 0.52
};
\addlegendentry{$CV_{0.025}$}
\addplot [semithick, mediumseagreen7117876, dashed, mark=square*, mark size=2, mark options={solid}]
table {%
0 0.13
1 0.13
2 0.26
3 0.51
4 0.52
};
\addlegendentry{$CV_{0.1}$}
\nextgroupplot[
tick align=outside,
tick pos=left,
x grid style={darkgray176},
xmin=-0.2, xmax=4.2,
xtick style={color=black},
xtick={0,1,2,3,4},
xticklabels={,,},
y grid style={darkgray176},
ylabel={crash rate},
ymin=0.028, ymax=0.732,
ytick style={color=black}
]
\addplot [semithick, peru17816071, line width=2pt]
table {%
0 0.05
1 0.06
2 0.11
3 0.06
4 0.09
};
\addplot [semithick, sienna1787171, mark=*, mark size=2, mark options={solid}]
table {%
0 0.06
1 0.06
2 0.08
3 0.08
4 0.12
};
\addplot [semithick, sienna1787171, dotted, mark=triangle*, mark size=2, mark options={solid}]
table {%
0 0.06
1 0.06
2 0.1
3 0.07
4 0.1
};
\addplot [semithick, sienna1787171, dashed, mark=square*, mark size=2, mark options={solid}]
table {%
0 0.07
1 0.07
2 0.08
3 0.1
4 0.1
};
\addplot [semithick, darkslateblue7671178, mark=*, mark size=2, mark options={solid}]
table {%
0 0.19
1 0.42
2 0.61
3 0.66
4 0.7
};
\addplot [semithick, darkslateblue7671178, dotted, mark=triangle*, mark size=2, mark options={solid}]
table {%
0 0.14
1 0.38
2 0.51
3 0.42
4 0.63
};
\addplot [semithick, darkslateblue7671178, dashed, mark=square*, mark size=2, mark options={solid}]
table {%
0 0.16
1 0.37
2 0.4
3 0.47
4 0.6
};
\addplot [semithick, mediumseagreen7117876, mark=*, mark size=2, mark options={solid}]
table {%
0 0.11
1 0.3
2 0.34
3 0.3
4 0.28
};
\addplot [semithick, mediumseagreen7117876, dotted, mark=triangle*, mark size=2, mark options={solid}]
table {%
0 0.07
1 0.11
2 0.27
3 0.24
4 0.27
};
\addplot [semithick, mediumseagreen7117876, dashed, mark=square*, mark size=2, mark options={solid}]
table {%
0 0.06
1 0.26
2 0.31
3 0.25
4 0.3
};
\nextgroupplot[
tick align=outside,
tick pos=left,
x grid style={darkgray176},
xmin=-0.2, xmax=4.2,
xtick style={color=black},
xtick={0,1,2,3,4},
y grid style={darkgray176},
xticklabels={0,50,100,200,300},
ylabel={wins-per-crash},
ymin=-0.115, ymax=9.6,
ytick style={color=black}
]
\addplot [semithick, peru17816071, line width=2pt]
table {%
0 2.2
1 3.5
2 3.363636
3 9.5
4 5.88888
};
\addplot [semithick, sienna1787171, mark=*, mark size=2, mark options={solid}]
table {%
0 1.83333333333333
1 2.33333333333333
2 5.125
3 6.875
4 4.08333333333333
};
\addplot [semithick, sienna1787171, dotted, mark=triangle*, mark size=2, mark options={solid}]
table {%
0 2
1 2.33333333333333
2 4
3 8
4 4.8
};
\addplot [semithick, sienna1787171, dashed, mark=square*, mark size=2, mark options={solid}]
table {%
0 1.71428571428571
1 1.71428571428571
2 5.125
3 5.1
4 4.7
};
\addplot [semithick, darkslateblue7671178, mark=*, mark size=2, mark options={solid}]
table {%
0 0.631578947368421
1 0.428571428571429
2 0.278688524590164
3 0.333333333333333
4 0.271428571428571
};
\addplot [semithick, darkslateblue7671178, dotted, mark=triangle*, mark size=2, mark options={solid}]
table {%
0 1.07142857142857
1 0.5
2 0.627450980392157
3 0.857142857142857
4 0.428571428571429
};
\addplot [semithick, darkslateblue7671178, dashed, mark=square*, mark size=2, mark options={solid}]
table {%
0 0.9375
1 0.567567567567568
2 0.825
3 0.702127659574468
4 0.516666666666667
};
\addplot [semithick, mediumseagreen7117876, mark=*, mark size=2, mark options={solid}]
table {%
0 1.27272727272727
1 0.633333333333333
2 0.970588235294118
3 1.63333333333333
4 1.96428571428571
};
\addplot [semithick, mediumseagreen7117876, dotted, mark=triangle*, mark size=2, mark options={solid}]
table {%
0 1.85714285714286
1 1.27272727272727
2 1
3 1.875
4 1.92592592592593
};
\addplot [semithick, mediumseagreen7117876, dashed, mark=square*, mark size=2, mark options={solid}]
table {%
0 2.16666666666667
1 0.5
2 0.838709677419355
3 2.04
4 1.73333333333333
};
\nextgroupplot[
tick align=outside,
tick pos=left,
x grid style={darkgray176},
xmin=-0.2, xmax=4.2,
xtick style={color=black},
xtick={0,1,2,3,4},
xticklabels={0,50,100,200,300},
y grid style={darkgray176},
ylabel={\(\displaystyle \text{min.} \ a_x\)},
ymin=-1.54834208180363, ymax=-0.278534966665986,
ytick style={color=black}
]
\addplot [semithick, peru17816071, line width=2pt]
table {%
0 -0.519452931243513
1 -0.9613496843426
2 -1.24394730712372
3 -0.875323352022802
4 -0.958106322526545
};
\addplot [semithick, sienna1787171, mark=*, mark size=2, mark options={solid}]
table {%
0 -0.448479906881927
1 -0.695101576235878
2 -0.727538346044256
3 -0.714430906355944
4 -0.871607733187644
};
\addplot [semithick, sienna1787171, dotted, mark=triangle*, mark size=2, mark options={solid}]
table {%
0 -0.419267103494459
1 -0.716251093734238
2 -0.763752801898546
3 -0.703845087320602
4 -0.82231587031762
};
\addplot [semithick, sienna1787171, dashed, mark=square*, mark size=2, mark options={solid}]
table {%
0 -0.42222114488103
1 -0.66521535126411
2 -0.807351452509593
3 -0.684345590863186
4 -0.860414672608281
};
\addplot [semithick, darkslateblue7671178, mark=*, mark size=2, mark options={solid}]
table {%
0 -0.376881236514632
1 -0.994485567670826
2 -1.29111224451146
3 -1.39705248026796
4 -1.4906235765701
};
\addplot [semithick, darkslateblue7671178, dotted, mark=triangle*, mark size=2, mark options={solid}]
table {%
0 -0.373741197766053
1 -0.723170205392903
2 -0.890394086547357
3 -1.11534585772843
4 -1.15483176979229
};
\addplot [semithick, darkslateblue7671178, dashed, mark=square*, mark size=2, mark options={solid}]
table {%
0 -0.336253471899515
1 -0.706564877656265
2 -0.938331451808784
3 -1.08336210404645
4 -1.25813344182015
};
\addplot [semithick, mediumseagreen7117876, mark=*, mark size=2, mark options={solid}]
table {%
0 -0.515959097817902
1 -1.09684952343133
2 -1.2634768959616
3 -1.18221351165042
4 -1.04549866181174
};
\addplot [semithick, mediumseagreen7117876, dotted, mark=triangle*, mark size=2, mark options={solid}]
table {%
0 -0.405996532195941
1 -1.04833445228062
2 -1.30517400819731
3 -1.13146899289305
4 -1.17571792454042
};
\addplot [semithick, mediumseagreen7117876, dashed, mark=square*, mark size=2, mark options={solid}]
table {%
0 -0.452971120200953
1 -1.07879226523913
2 -1.32936370686126
3 -1.00889660964955
4 -1.16949412801012
};
\end{groupplot}
% \vspace{-0.1cm}
%\node[below = 0.5cm of eval c2r4.south] {(b) Inside};
%\node[below = 0.5cm of eval c1r4.south] {(a) Inside-outside};
\end{tikzpicture} 
    \vspace{-0.6cm}
    \caption{Monte Carlo closed-loop performance comparison between the studied prediction models. The $x$ axis corresponds to the values of $q_y^{(2)}$, the weight on the TV blocking cost. Higher values correspond to more aggressive blocking behavior by the TV during a race. The subscripts denote the values used for the safety bound parameters. For GP, it corresponds to the number of standard deviations $\gamma$. For CV and NL, it corresponds to the radius of the circular safety bound in meters. The grey line (GT) indicates the best performance achieved by using the ground truth open-loop solutions in lieu of predictions.} 
    \label{fig:clmetrics}
    \vspace{-0.3cm}
\end{figure*}
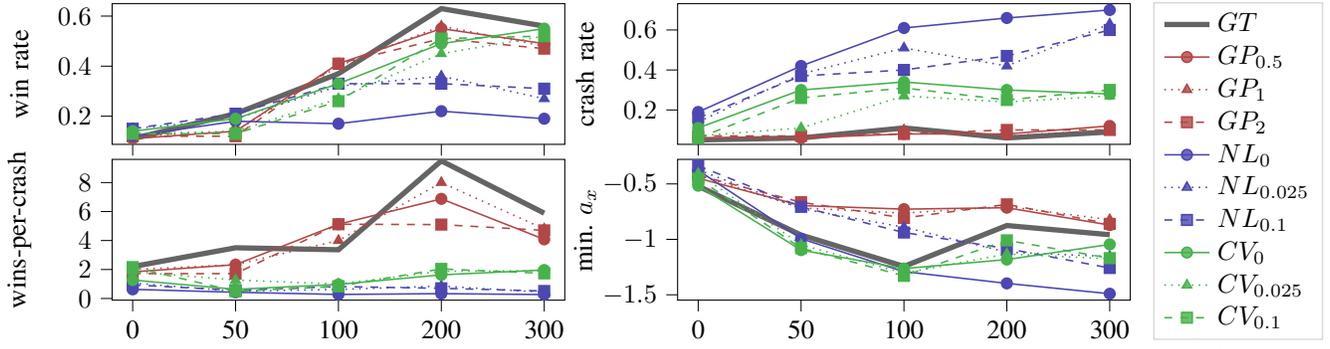

Next, the closed-loop performance in the presence of TV model mismatch is evaluated. To simulate model mismatch, we vary the blocking aggressiveness $q_y^{(2)}$ of the TV policy while keeping the predictors unchanged, i.e. no retraining or additional tuning is done. As a performance baseline, we include race results where the open-loop solutions of the TV policy are used in lieu of a prediction when computing the control action of the EV. This corresponds to the case where exact knowledge of the TV's future plans are known to the EV, we call this the ground truth (GT) case. For each class of predictor, we additionally vary the size of the safety bound to investigate its effect on closed-loop race performance. For the GP, we pick safety bounds with $\gamma = 0.5, 1, 2$. For the CAV and NL predictors, circular safety bounds of radius 0.025, and 0.1 m are added. These are chosen such that the safety bound size on average is similar to the safety bounds induced by the GP's uncertainties in the one and two standard deviation cases. The performance of the EV policy in closed-loop with each of these predictors against the TV blocking policy with the various aggressiveness factor settings are depicted in Fig.~\ref{fig:clmetrics}.

In terms of win rate, shown in the upper left, GP clearly outperforms the other predictors for the TV policy that it was trained on, i.e. $q_y^{(2)}=200$. For the other settings of $q_y^{(2)}$, while GP does not impart a clear advantage in win rate, it remains competitive with the other predictors, including the GT case when we use the TV open-loop solutions directly instead of the output of a predictor. In addition, GP seems to be less sensitive against variations in the size of the safety bound, likely because it provides consistently accurate predictions and does not have to rely on large safety bounds to keep the vehicle safe. In general, all predictors are able to exploit overly aggressive behavior by the target vehicle by overtaking through the space which is made free from aggressive blocking attempts. This is clearly seen in the results as the win rate of the ego vehicle increases with the aggressiveness of the target vehicle. 

As for the crash rate, shown in the upper right, GP is able to maintain a crash rate below $10\%$ for all values of $q_y^{(2)}$, whereas the other predictors cannot keep the EV safe when the TV exhibits more aggressive blocking behavior. In fact, it is clear from this plot that in terms of safety, GP is essentially able to achieve identical performance to GT where the TV's future plans are known to the EV. Generally, we observe that GP outperforms the other predictors in terms of safety as it is able to accurately anticipate the TV's blocking attempts and react accordingly by braking or steering out of the way, ultimately reducing the number of crashes. 

While the ultimate objective of a race is to win, it is also worthwhile to look at the the outcome of races where the EV loses to the TV. In particular, we are interested in whether the EV was able to finish the race safely despite the loss or if the loss was caused by a crash. The reason for this concerns the practical aspects of car racing where crashes can be catastrophic and expensive. As such, consider two predictors A and B where Predictor A has a win rate of 40\% and a crash rate of 5\% and Predictor B has a win rate of 50\% and a crash rate of 20\%. Despite the lower win rate, it would be prudent to prefer Predictor A over Predictor B due to its significantly lower crash rate. In order to quantify this preference, we introduce the metric wins-per-crash, to normalize for the effect of crashing in the win rate and to measure the trade-off between wins and safety. A high score for this metric requires that the predictor not only lead to more wins but also few crashes. The results for this metric are shown in the lower left, where it is clear that GP outperforms the other predictors across all TV policies. We additionally see that GP closely tracks the results from using GT.

We finally examine the largest deceleration (averaged over all races) experienced by the EV under each predictor, which is shown in the plot in the lower right. This acts as a proxy to illustrate how well the predictors anticipate the TV's behavior. If blocking attempts are predicted accurately, we expect the EV to be able to react in such a way that minimizes acceleration, which is encouraged by the input cost in \eqref{eq:mpcc_cost}. On the other hand, if blocking maneuvers by the TV are not predicted, we expect the EV to be more reactionary to the current state of the TV, which would lead to larger throttle and braking commands. From the plot, it can be clearly seen that under NL and CV, the EV generally experiences larger decelerations, with more extreme values being observed for more aggressive TV behavior. On the other hand, GP is less sensitive to the target vehicle's aggressiveness and experiences less extreme deceleration indicating that it is able to anticipate the incoming blocking attempts by the TV and can begin to react earlier. We note that in this aspect, GP actually outperforms GT. We believe that this is due to the fact that GP was trained on closed-loop TV behavior and is predicting the \emph{actual} trajectory which will be driven by the TV. On the other hand, the open-loop predictions from the TV policies, which are used in GT, can change drastically between consecutive time steps due to local optimality of NLPs. This can lead to significant mismatch between the open-loop predictions and the \emph{actual} driven trajectory, which ultimately results in larger braking commands to react to those mismatches.

% Overall, the Monte Carlo study shows that when using the GP-based predictor in closed-loop with the MPC policy, the EV is able to better balance performance and safety. 

\subsection{Hardware Study}
% \begin{figure*}[t]
%      \centering
%      \input{plots/nlmpcovertake}
%      \input{plots/gpovertake}
% \end{figure*}
\begin{figure*}[t]
     \centering
     \includegraphics[trim={0.1cm 0 0.1cm 0},clip,width=0.99\textwidth]{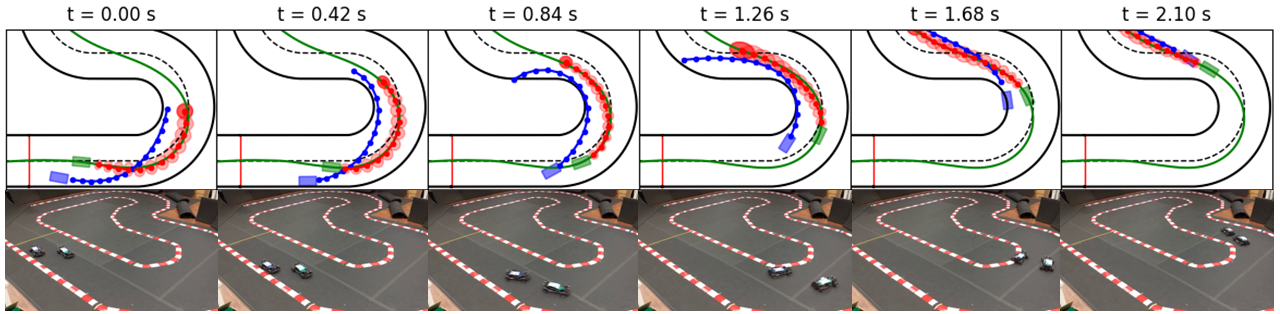}
     \includegraphics[trim={0.1cm 0 0 0.6cm},clip,width=0.99\textwidth]{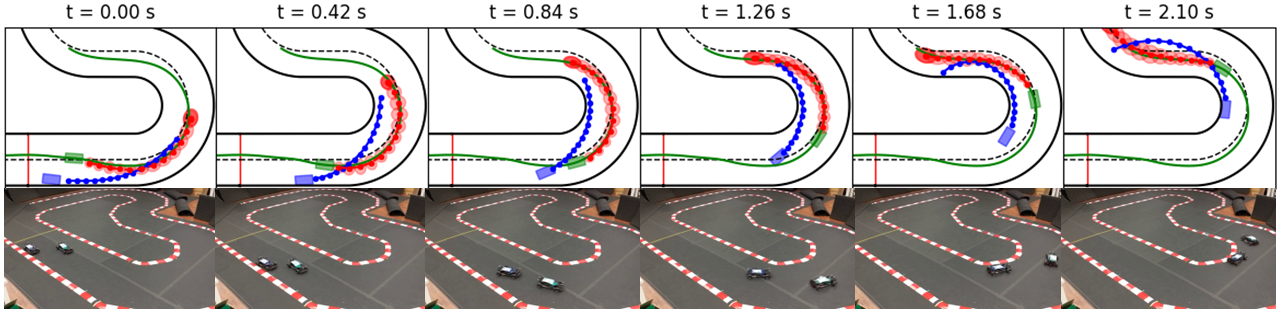}
     \includegraphics[trim={0 0 0.1cm 0.6cm},clip,width=0.99\textwidth]{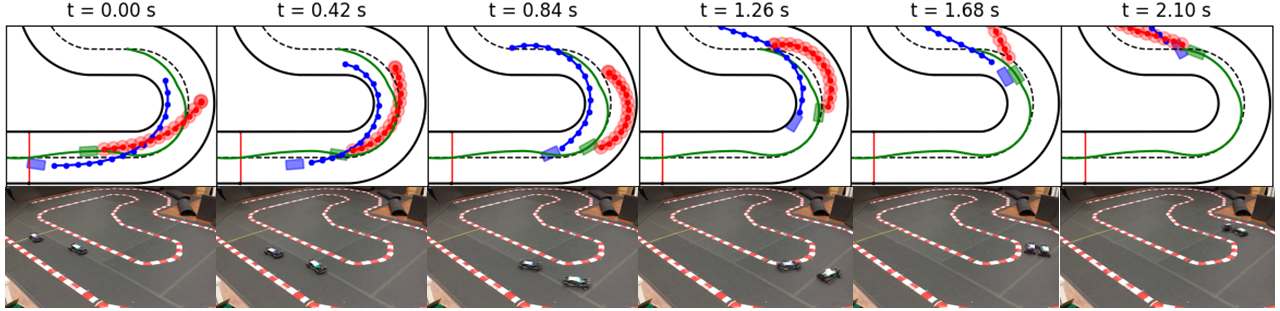}
     \vspace{-0.1cm}
     \caption{EV (blue) overtaking attempts against the TV (green) with predictions in red at six time instances during the hardware experiment. The first two rows show successful and failed attempts with GP and the third row shows a failed attempt with NL. The red ellipses and circles correspond to the uncertainty-expanded or artificial safety bounds around the nominal prediction at each step in the MPC horizon. The solid green line indicates the closed-loop trajectory of TV. Video recordings of all hardware experiments can be found at \url{https://youtu.be/KMSs4ofDfIs}.}
     \label{fig:overtake_compare}
     \vspace{-0.7cm}
\end{figure*}

\begin{table}[t]
\centering
\begin{tabular}{ c|c|c } 
 & GP & NL \\
\midrule[1pt]
Overtake & 4 & 3 \\ 
Safe & 5 & 0 \\
Collision (minor) & 1 & 2 \\
Collision (major) & 0 & 5 \\
\end{tabular}
\caption{Race outcomes}
\label{tab:outcomes}
% \vspace{-0.7cm}
\end{table}

In our hardware experiments, we compare the performance of the EV MPCC policy in closed-loop with the NL and GP predictors in ten races on the L-shaped track shown in Fig.~\ref{fig:overtake_compare}. We chose $\gamma=1$ for GP and a radius of 0.14 m for NL such that the safety bound for NL is similar in size to those produced by GP. Using each of the two predictors, we run ten races with the EV starting at the same position. The races end after the EV has driven three laps or after a major collision occurs. As for the TV, we chose different starting positions ahead of the EV for each race. These starting positions are kept constant in the corresponding race where GP and NL predictions are used. For all experiments, we impose a longitudinal speed constraint of 2.8 and 2.0 m/s for the EV and TV respectively. This is done to ensure that close interactions will occur between the vehicles during each race. As in the simulation study, the TV uses the blocking policy described in Section~\ref{sec:data_gen} whereas the EV is encouraged to overtake the TV through a higher weight on the progress maximization cost in \eqref{eq:mpcc_cost}. We note that both control policies use a horizon length of $N=15$ and no additional controller tuning was performed when switching between the GP and NL predictors. The outcomes of the ten races with the GP and NL predictors is summarized in Table~\ref{tab:outcomes}, where the EV was able to perform safe overtaking maneuvers (all with longitudinal velocity greater than 2.7 m/s) and win in four races with GP and in three races with NL. For the remaining races, the EV lost but was able to remain safe in five races with GP and saw only one minor collision where the wheels of the two vehicles touched. On the other hand, the TV saw collisions in all seven remaining races. Out of these collisions, two were minor wheel touches, but the other five were major collisions which significantly altered the trajectory of both cars. We note that these outcomes are in close agreement with the results from our simulation study.

\begin{figure}[t]
     \centering
     \includegraphics[width=0.90\columnwidth]{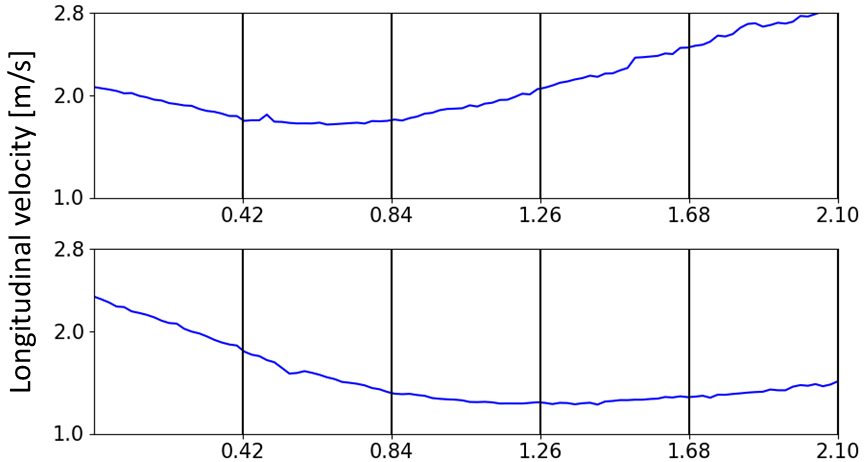}
     \vspace{-0.1cm}
     \caption{Longitudinal velocity profiles for the successful (top) and failed (bottom) overtaking attempts using GP. The black vertical lines correspond to the time instances of the frames in Fig.~\ref{fig:overtake_compare}.}
     \label{fig:speed_compare}
     % \vspace{-0.5cm}
\end{figure}

To further understand the advantages that GP affords us when compared to NL, we turn to a similar interaction which arose in multiple instances with both predictors, which is illustrated in Fig.~\ref{fig:overtake_compare}. The scenario begins with EV on the outside of TV and the two cars approaching the first 180\degree \ turn. The top frames show the case when GP is used where a successful inside overtake is performed. In this interaction, it can be seen that the TV's closed-loop blocking maneuver is contained in the uncertainty-expanded safety bounds. By accurately predicting the behavior of the TV and recognizing that it cannot successfully block the EV, this allows the EV to exploit the free space on the inside of the TV to perform a safe overtake. The middle frames show the case when GP is used, but the EV was not able to successfully overtake the TV. However, it is able to remain safe and avoid collisions. In this scenario, the EV begins by slowing down in anticipation of the TV's outside blocking maneuver as seen in Fig.~\ref{fig:speed_compare}. This lower speed then prevents the EV from finding a safe solution for performing an inside overtake as it accurately predicts that the TV will be able to block it even as it takes the inside line. The EV therefore slows down and remains behind the TV. Finally, the bottom frames show the case when NL is used and the EV collides with the TV during its overtaking attempt. This is primarily due to the inaccurate predictions from NL, which do not take into account the blocking incentive of the TV, which, as seen in \eqref{eq:blocking_cost}, dominates the TV's MPC cost especially when the two cars are in close proximity with each other. The added safety bounds are of no use here either due to the significant divergence of the NL predictions from the closed-loop trajectory of the TV.

\section{Conclusions and Future Work}
In this work, we presented a learning-based method for predicting opponent behavior in the context of head-to-head competitive autonomous racing. In particular, Gaussian processes were trained on data from past races to generate trajectory predictions and their corresponding uncertainties, which are then used in closed-loop with an MPCC policy. We show through both simulation and hardware experiments that our approach outperforms non-data-driven predictors in terms of safety while maintaining a high win rate.
Future work will focus on an extension to an arbitrary number of opponents and online fine-tuning of the predictor, which would allow for adaptation to previously unseen behavior.

%%%%%%%%%%%%%%%%%%%%%%%%%%%%%%%%%%%%%%%%%%%%%%%%%%%%%%%%%%%%%%%%%%%%%%%%%%%%%%%%

\bibliographystyle{IEEEtran}
\bibliography{main.bib}

% Generated by IEEEtran.bst, version: 1.14 (2015/08/26)
\begin{thebibliography}{10}
\providecommand{\url}[1]{#1}
\csname url@samestyle\endcsname
\providecommand{\newblock}{\relax}
\providecommand{\bibinfo}[2]{#2}
\providecommand{\BIBentrySTDinterwordspacing}{\spaceskip=0pt\relax}
\providecommand{\BIBentryALTinterwordstretchfactor}{4}
\providecommand{\BIBentryALTinterwordspacing}{\spaceskip=\fontdimen2\font plus
\BIBentryALTinterwordstretchfactor\fontdimen3\font minus
  \fontdimen4\font\relax}
\providecommand{\BIBforeignlanguage}[2]{{%
\expandafter\ifx\csname l@#1\endcsname\relax
\typeout{** WARNING: IEEEtran.bst: No hyphenation pattern has been}%
\typeout{** loaded for the language `#1'. Using the pattern for}%
\typeout{** the default language instead.}%
\else
\language=\csname l@#1\endcsname
\fi
#2}}
\providecommand{\BIBdecl}{\relax}
\BIBdecl

\bibitem{hong2019rules}
J.~Hong, B.~Sapp, and J.~Philbin, ``Rules of the road: Predicting driving
  behavior with a convolutional model of semantic interactions,'' \emph{2019
  IEEE/CVF Conference on Computer Vision and Pattern Recognition (CVPR)}, pp.
  8446--8454, 2019.

\bibitem{djuric2020uncertaintyaware}
N.~Djuric, V.~Radosavljevic, H.~Cui, T.~Nguyen, F.-C. Chou, T.-H. Lin,
  N.~Singh, and J.~G. Schneider, ``Uncertainty-aware short-term motion
  prediction of traffic actors for autonomous driving,'' \emph{2020 IEEE Winter
  Conference on Applications of Computer Vision (WACV)}, pp. 2084--2093, 2020.

\bibitem{WaymoZhao2020TNTTT}
H.~Zhao, J.~Gao, T.~Lan, C.~Sun, B.~Sapp, B.~Varadarajan, Y.~Shen, Y.~Shen,
  Y.~Chai, C.~Schmid, C.~Li, and D.~Anguelov, ``Tnt: Target-driven trajectory
  prediction,'' in \emph{CoRL}, 2020.

\bibitem{Salzmann2020TrajectronDT}
T.~Salzmann, B.~Ivanovic, P.~Chakravarty, and M.~Pavone, ``Trajectron++:
  Dynamically-feasible trajectory forecasting with heterogeneous data,'' in
  \emph{ECCV}, 2020.

\bibitem{UberCui2020DeepKM}
H.~Cui, T.~Nguyen, F.-C. Chou, T.-H. Lin, J.~G. Schneider, D.~Bradley, and
  N.~Djuric, ``Deep kinematic models for kinematically feasible vehicle
  trajectory predictions,'' \emph{2020 IEEE International Conference on
  Robotics and Automation (ICRA)}, pp. 10\,563--10\,569, 2020.

\bibitem{Vallon2022DataDrivenSF}
C.~Vallon and F.~Borrelli, ``Data-driven strategies for hierarchical predictive
  control in unknown environments,'' \emph{ArXiv}, vol. abs/2105.06005, 2022.

\bibitem{lefevre2015driver}
S.~Lef{\`e}vre, A.~Carvalho, Y.~Gao, H.~E. Tseng, and F.~Borrelli, ``Driver
  models for personalised driving assistance,'' \emph{Vehicle System Dynamics},
  vol.~53, no.~12, pp. 1705--1720, 2015.

\bibitem{Yoon2021InteractionAwarePT}
Y.~Yoon, C.~Kim, J.~Lee, and K.~Yi, ``Interaction-aware probabilistic
  trajectory prediction of cut-in vehicles using gaussian process for proactive
  control of autonomous vehicles,'' \emph{IEEE Access}, vol.~9, pp.
  63\,440--63\,455, 2021.

\bibitem{Brdigam2021GaussianPS}
T.~Br{\"u}digam, A.~Capone, S.~Hirche, D.~Wollherr, and M.~Leibold, ``Gaussian
  process-based stochastic model predictive control for overtaking in
  autonomous racing,'' \emph{ArXiv}, vol. abs/2105.12236, 2021.

\bibitem{Kong2015}
J.~Kong, M.~Pfeiffer, G.~Schildbach, and F.~Borrelli, ``{Kinematic and dynamic
  vehicle models for autonomous driving control design},'' \emph{IEEE
  Intelligent Vehicles Symposium, Proceedings}, vol. 2015-August, pp.
  1094--1099, 2015.

\bibitem{micaelli1993trajectory}
A.~Micaelli and C.~Samson, ``Trajectory tracking for unicycle-type and
  two-steering-wheels mobile robots,'' Ph.D. dissertation, INRIA, 1993.

\bibitem{liniger2015optimization}
A.~Liniger, A.~Domahidi, and M.~Morari, ``Optimization-based autonomous racing
  of 1: 43 scale rc cars,'' \emph{Optimal Control Applications and Methods},
  vol.~36, no.~5, pp. 628--647, 2015.

\bibitem{williams2006gaussian}
C.~K. Williams and C.~E. Rasmussen, \emph{Gaussian processes for machine
  learning}.\hskip 1em plus 0.5em minus 0.4em\relax MIT press Cambridge, MA,
  2006.

\bibitem{lehtonen2013look}
E.~Lehtonen, O.~Lappi, H.~Kotkanen, and H.~Summala, ``Look-ahead fixations in
  curve driving,'' \emph{Ergonomics}, vol.~56, no.~1, pp. 34--44, 2013.

\bibitem{gardner2018gpytorch}
J.~Gardner, G.~Pleiss, K.~Q. Weinberger, D.~Bindel, and A.~G. Wilson,
  ``Gpytorch: Blackbox matrix-matrix gaussian process inference with gpu
  acceleration,'' \emph{Advances in neural information processing systems},
  vol.~31, 2018.

\bibitem{Hensman2015ScalableVG}
J.~Hensman, A.~G. de~G.~Matthews, and Z.~Ghahramani, ``Scalable variational
  gaussian process classification,'' in \emph{AISTATS}, 2015.

\bibitem{Schwarting2017ParallelAI}
W.~Schwarting, J.~Alonso-Mora, L.~Paull, S.~Karaman, and D.~Rus, ``Parallel
  autonomy in automated vehicles: Safe motion generation with minimal
  intervention,'' \emph{2017 IEEE International Conference on Robotics and
  Automation (ICRA)}, pp. 1928--1935, 2017.

\bibitem{Andersson2019}
J.~A.~E. Andersson, J.~Gillis, G.~Horn, J.~B. Rawlings, and M.~Diehl,
  ``{CasADi} -- {A} software framework for nonlinear optimization and optimal
  control,'' \emph{Mathematical Programming Computation}, vol.~11, no.~1, pp.
  1--36, 2019.

\bibitem{frison2020hpipm}
G.~Frison and M.~Diehl, ``Hpipm: a high-performance quadratic programming
  framework for model predictive control,'' \emph{IFAC-PapersOnLine}, vol.~53,
  no.~2, pp. 6563--6569, 2020.

\end{thebibliography}

\end{document}